\documentclass{article}

\usepackage{microtype}
\usepackage{graphicx}
\usepackage{subcaption}
\usepackage{booktabs} 

\newcommand{\SeqLen}{N} 

\newcommand{\PrefillLen}{N_{\text{pre}}} 

\usepackage[preprint]{icml2026}

\usepackage{amsmath}
\usepackage{amssymb}
\usepackage{mathtools}
\usepackage{amsthm}

\usepackage[capitalize,noabbrev]{cleveref}

\theoremstyle{plain}

\theoremstyle{definition}

\theoremstyle{remark}

\usepackage[textsize=tiny]{todonotes}

\icmltitlerunning{CSAttention: Centroid-Scoring Attention for Accelerating LLM Inference}

\begin{document}

\twocolumn[
  \icmltitle{CSAttention: Centroid-Scoring Attention for Accelerating LLM Inference}



  \icmlsetsymbol{equal}{*}

  \begin{icmlauthorlist}
    \icmlauthor{Chuxu Song}{rutgers}
    \icmlauthor{Zhencan Peng}{rutgers}
    \icmlauthor{Jiuqi Wei}{ant}
    \icmlauthor{Chuanhui Yang}{ant}
  \end{icmlauthorlist}

  \icmlaffiliation{rutgers}{Department of Computer Science, Rutgers University}
  \icmlaffiliation{ant}{Oceanbase, Ant Group}

  \icmlcorrespondingauthor{Jiuqi Wei}{weijiuqi.wjq@antgroup.com}

  \icmlkeywords{Machine Learning, ICML}

  \vskip 0.3in
]



\printAffiliationsAndNotice{}  

\begin{abstract}

Long-context LLMs increasingly rely on extended, reusable prefill prompts for agents and domain Q\&A, pushing attention and KV-cache to become the dominant decode-time bottlenecks. 
While sparse attention reduces computation and transfer costs, it often struggles to maintain accuracy at high sparsity levels due to the inherent distribution shift between Queries and Keys.
We propose Centroid-Scoring Attention (CSAttention), a training-free sparse attention method optimized for high-throughput serving of reusable contexts.
CSAttention adopts a \emph{storage-for-computation} strategy tailored to the offline-prefill / online-decode setting: it front-loads computation into a one-time offline prefill phase that can be amortized across multiple queries, while aggressively optimizing per-step decoding latency.
Specifically, CSAttention constructs query-centric lookup tables during offline prefill, whose size remains fixed during decoding, and enables online decoding to replace full-context scans with efficient table lookups and GPU-friendly score accumulation.
Extensive experiments demonstrate that CSAttention achieves near-identical accuracy to full attention. 
Under high sparsity (95\%) and long-context settings (32K–128K), CSAttention consistently outperforms state-of-the-art sparse attention methods in both model accuracy and inference speed, achieving up to 4.6$\times$ inference speedup over the most accurate baseline at a context length of 128K.



\end{abstract}

\section{Introduction}

Long-context LLM usage is rapidly accelerating, driven by LLM agents and domain-specific Q\&A workflows that require \emph{very long, reusable prompts}.
In many real-world deployments, these workloads naturally decompose into an \emph{offline prefill} stage and an \emph{online decode} stage~\citep{lu2024turborag, jin2024ragcache, gao2024cost, lee2025shared}.
This separation is most evident in ``Write-Once, Read-Many'' serving scenarios—such as context caching for RAG pipelines or agent schemas—where servers execute a one-time prefill over a long context of length $\PrefillLen$ to materialize reusable KV states and auxiliary search structures, persisting them outside HBM (e.g., in CPU DRAM or SSD) for repeated reuse.
At request time, online decoding proceeds over the growing context of length $\SeqLen{=}\PrefillLen{+}t$, loading the required artifacts on demand and performing attention for each decoding step, while allowing only lightweight maintenance (e.g., appending new keys) to preserve predictable latency.
Under this serving model, the prefill phase may process tens of thousands to millions of tokens and amortize its cost across prolonged generation or repeated requests, making \emph{per-step decoding latency and throughput under long contexts} the primary optimization target.



While the quadratic computation of dense attention is a classical concern, in practical long-context serving the dominant bottleneck during online decoding is often the KV cache.
Its memory footprint scales linearly with context length, model dimension, and attention heads, quickly saturating GPU HBM and forcing systems to page KV states between GPU and CPU memory, incurring substantial bandwidth and latency overheads.
Modern serving stacks, such as vLLM with PagedAttention, mitigate fragmentation and improve KV sharing, but do not change the fundamental scaling behavior~\citep{kwon2023pagedattention}.
Even with grouped-query or multi-query attention (GQA/MQA) reducing the number of KV heads~\citep{shazeer2019mqa, ainslie2023gqa}, long contexts remain challenging: for example, Llama-3.1-8B at \mbox{1M} tokens requires on the order of $10^2$,GB of KV memory in bf16 without approximation, far exceeding a single GPU’s capacity~\citep{luo2025headinfer}.

A natural response is sparse attention: standard attention matrices exhibit inherent sparsity, wherein a large fraction of the computed weights are close to zero and can be pruned without significant impact on output quality~\citep{zhang2025efficient}.
Therefore, the model can reliably attend to only a small fraction of keys, reducing attention FLOPs and the effective KV touched per step simultaneously.
Prior work explores three main directions.
(i) Token eviction/retention: dynamically keep only ``heavy-hitter'' tokens in the cache (e.g., H$_2$O and follow-ups)~\citep{zhang2023h2o,qhitter2024}, which prunes storage but can be sensitive to online prediction errors.
(ii) Bandwidth-aware fetching: techniques like SparQ selectively fetch historical KV to raise memory-bandwidth efficiency during attention~\citep{ribar2024sparq}.
(iii) Index-based retrieval: treat KV search as MIPS/ANN over quantized representations (e.g., PQCache) or use sampling via LSH (e.g., MagicPig) to approximate attention~\citep{zhang2025pqcache,chen2024magicpig}.
While these methods reduce computational and transfer costs, they encounter a fundamental \emph{challenge at high sparsity}:
existing indexes often rely on Key-based clustering, which suffers from Query-Key distribution shift, making it exceedingly difficult to simultaneously maintain high model performance and achieve fast inference speed.
Moreover, at long-context serving time, even a modest recall drop can quickly compound into quality degradation, so robustness under distribution drift (e.g., long-generation) becomes essential.

In this work, we propose Centroid-Scoring Attention (CSAttention), a training-free sparse attention method that accelerates long-context serving.
To achieve high model performance and fast inference under high sparsity, CSAttention adopts a \emph{storage-for-computation} strategy: it leverages query distributions to construct query-centric lookup tables during the offline prefill stage, enabling online decoding to perform efficient searches and centroid-score accumulation over regular, GPU-friendly data structures.
This design explicitly trades a one-time $\PrefillLen$ offline build (acceptable under reusable prefill) for \emph{bounded-work} online decoding that avoids any full-context scans.
Specifically,
\begin{itemize}
\item \textbf{Query-centric tables (offline).} Split the feature space into $m$ subspaces.
For each subspace, cluster queries from prefill into $C$ centroids.
For every centroid, precompute partial dot-products with all keys in that subspace as centroid-scores, and store a compressed (bounded-capacity) Top-$L$ list (indices + scores).
This design amortizes the one-time indexing cost across many requests that share the same long prefill context.
\item \textbf{Keys retrieval (online).} For a new query, select its nearest centroid in each subspace (1-of-$C$ per subspace), fetch the $m$ short lists, and sum partial scores by key index on GPU.
Keys truly aligned with the query tend to exhibit high centroid-scores across multiple subspaces, rising to the top after sparse accumulation—mitigating potential query drift via subspace aggregation without scanning the whole cache.
As decoding streams in new keys, we \emph{incrementally} maintain these lists by attempting to insert each arriving key into the corresponding centroids’ Top-$L$ (only if its partial score exceeds the current minimum), keeping table capacity bounded and update overhead negligible.
\end{itemize}

\noindent \textbf{Why this helps.}
(i) By exploiting query-centric clustering offline, the index structure tracks the geometry of queries $Q$ rather than only keys $K$, mitigating $Q/K$ distribution shift and ensuring robustness against query drift during generation.
This shifts the retrieval structure from key-centric clustering to a query-centric serving primitive, which is crucial for sustaining recall at very high sparsity.
(ii) By utilizing compressed lookup tables and only running a small number of regular GPU kernels during decoding, CSAttention avoids per-query score movement and irregular control flow, sustaining high hardware utilization and inference speed.
(iii) By employing subspace partitioning and query-centric tables, CSAttention effectively recovers high-scoring keys under very high sparsity (e.g., 95\%), enabling significant computational savings while maintaining model accuracy.

\paragraph{Results at a glance.}
(i) \emph{Near-lossless accuracy at 95\% sparsity}: on LongBench evaluations across three models (Llama-3.1-8B, Qwen3-8B, and Mistral-7B), CSAttention maintains nearly identical accuracy to full attention (within 0.7\% loss) at 95\% sparsity.
(ii) \emph{Best accuracy and speed over competitors}:
under high sparsity (95\%) and long-context settings (32K–128K), CSAttention consistently outperforms state-of-the-art sparse attention methods in both model accuracy and inference speed, achieving up to 4.6$\times$ inference speedup over the most accurate baseline at a context length of 128K.
(iii) \emph{Robustness in Long-Generation}: On LongBench-v2 with Chain-of-Thought (CoT) decoding, CSAttention preserves accuracy even as the generation length extends, validating the stability of query-centric clusters against distribution drift.

\paragraph{Clarification: KV cache vs.\ Lookup tables.}
In this paper, the \emph{KV cache} refers to the standard storage of keys and values for all $\SeqLen$ context tokens, whose memory footprint necessarily scales linearly with context length.
In contrast, the \emph{lookup tables} introduced by CSAttention are \emph{auxiliary, bounded-capacity} data structures (Top-$L$ per centroid) constructed during offline prefill. 
Crucially, their capacity is \emph{fixed after prefill} and does not grow with the number of decoding steps $t$, even though the KV cache itself continues to scale with $\SeqLen{=}\PrefillLen{+}t$ during online decoding.
Accordingly, CSAttention does not seek to eliminate the linear KV storage cost; rather, it targets the dominant \emph{decode-time} compute/transfer bottleneck by ensuring that each decoding step touches only a small fraction of the KV cache with predictable online work.

\section{Observations and Motivation}

\begin{figure}[t]
    \centering
    \begin{subfigure}[t]{0.24\textwidth}
        \centering
        \includegraphics[width=\linewidth]{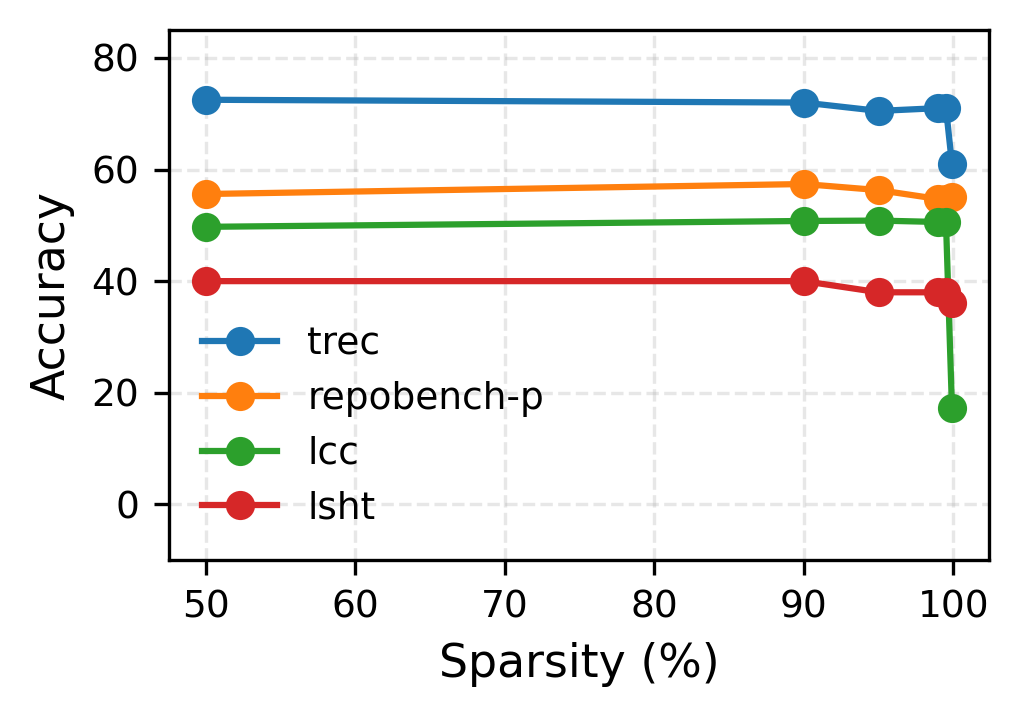}
        \caption{Accuracy vs. sparsity}
        \label{fig:SparsityAblation}
    \end{subfigure}\hfill
    \begin{subfigure}[t]{0.24\textwidth}
        \centering
        \includegraphics[width=\linewidth]{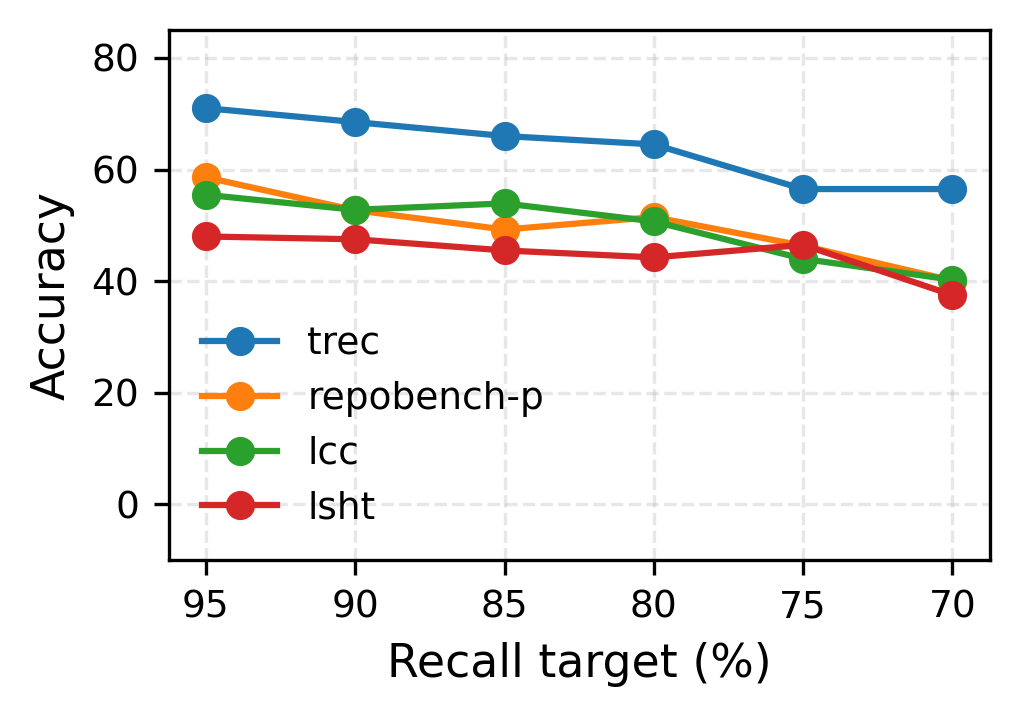}
        \caption{Accuracy vs. recall}
        \label{fig:RecallAblation}
    \end{subfigure}\hfill
    \begin{subfigure}[t]{0.24\textwidth}
        \centering
        \includegraphics[width=\linewidth]{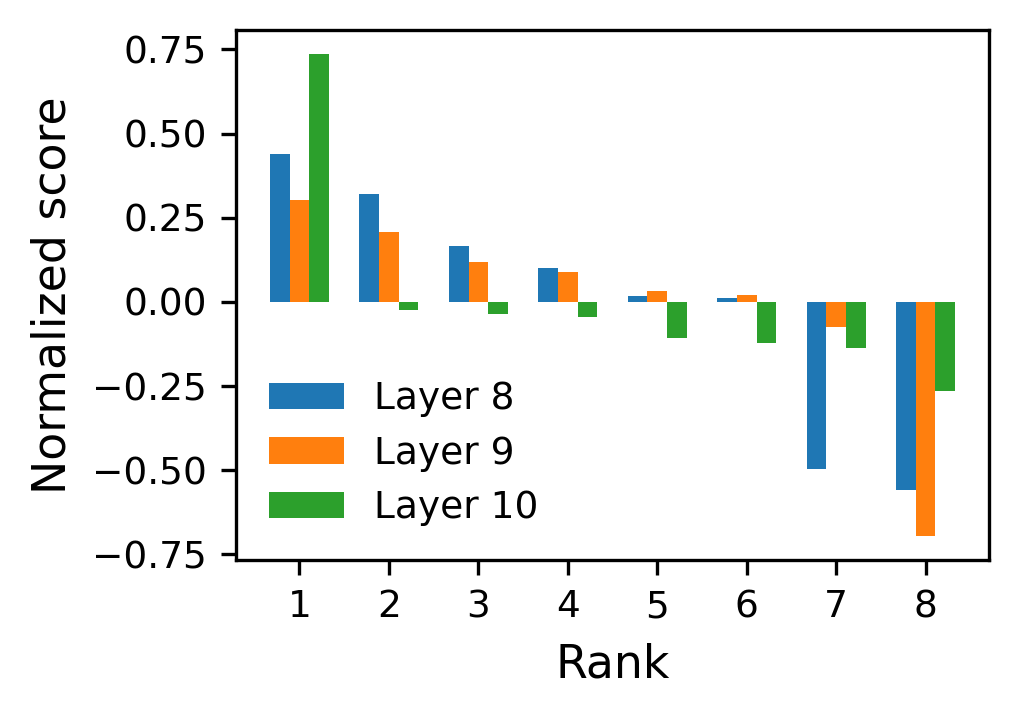}
        \caption{RankShares L8--10}
        \label{fig:RankSharesL8to10}
    \end{subfigure}\hfill
    \begin{subfigure}[t]{0.24\textwidth}
        \centering
        \includegraphics[width=\linewidth]{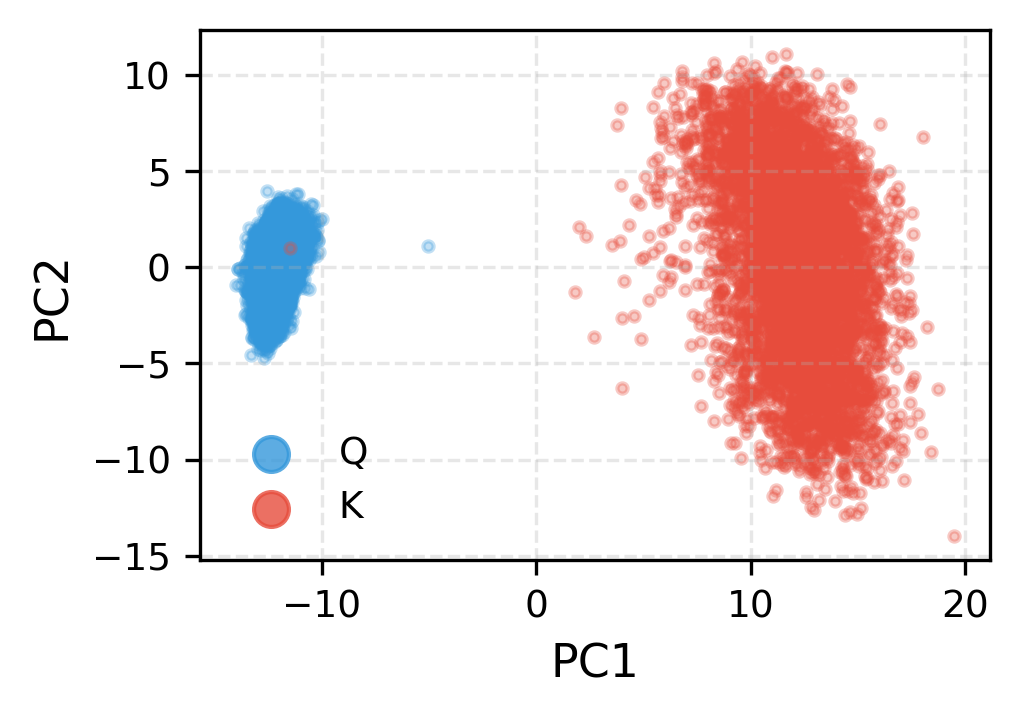}
        \caption{Q/K PCA (L10, H0)}
        \label{fig:OOD_PCA}
    \end{subfigure}\hfill
    \caption{
    \textbf{Observations.}
    (a/b) Accuracy vs. sparsity/recall on four LongBench tasks (Llama-3.1-8B-Instruct)
    (c) Heterogeneous subspace: rank-share of accumulated \(q\!\cdot\!k\) contributions across \(m=8\) subspaces, grouped over Layers 8--10.
    (d) PCA of queries $Q$ and keys $K$ from Llama-3.1-8B-Instruct (Layer 10, Head 0) shows a distribution shift between $Q$ and $K$.}\vspace{-1.0em}
    \label{fig:three-across}
\end{figure}


\subsection{Preliminaries and empirical sparsity of attention}
\label{sec:prelim}
Let $Q,K,V \in \mathbb{R}^{N \times d}$ be the query, key, and value sequences for one head, where
$d$ is the head dimension and $N$ is the \emph{current} (growing) context length during decoding.
At the current decode step, we denote the query vector by $q \in \mathbb{R}^{1 \times d}$, i.e., $q$ is a row of $Q$.
Scaled dot-product attention computes per-query weights
\[
a = \mathrm{softmax}\!\left(\frac{q K^\top}{\sqrt{d}}\right) \in \mathbb{R}^{1 \times N},
\]
and returns the weighted value aggregation $o = a V \in \mathbb{R}^{1 \times d}$.
The vector $a$ (the \emph{attention scores}) sums to 1 and determines which past tokens’
values contribute to the output. In multi-head attention (MHA/GQA/MQA), this is applied
per head and concatenated or averaged across heads.

Attention matrices in long-context LLMs are effectively sparse: most weights are near zero and only a small fraction of keys carry significant mass.
This suggests substantial headroom to prune keys and reduce compute/bandwidth, provided we reliably keep the truly high-weight keys~\citep{liu2024retrievalattention, zhang2025efficient}.

\subsection{Top-K recall governs accuracy in high sparsity}

Even modest misses in the true Top-K hurt accuracy. Since the true Top-K under dense attention is unavailable at decode time without full attention, any sparse-attention method must rely on an approximation.
Figure~\ref{fig:three-across}(b) shows accuracy strongly correlates with Top-K recall, especially at very high sparsity.
This indicates that preserving Top-K (or near-Top-K) keys is the governing factor for accuracy in extreme pruning regimes.
Thus, a practical sparse attention design should prioritize stable Top-K recall under long-context decoding.

\subsection{Importance deviation of different subspaces} \label{sec:subspace}
Search-based sparse attention methods essentially transform the attention retrieval step into a vector similarity search task~\citep{liu2024retrievalattention}.
Partitioning the $d$-dimensional origin space into $m$ $d/m$-dimensional subspaces and leveraging clustering to construct an auxiliary lookup structure in each subspace is a highly effective scheme in the field of vector similarity search~\citep{jegou2010product,wei2025subspace}.
The underlying principle is that from the perspective of accuracy, different subspaces contribute unevenly to the final $q\!\cdot\!k$ similarity.
Figure~\ref{fig:three-across}(c) shows that the accumulated $q\!\cdot\!k$ contributions exhibit a heavy-tailed distribution across subspaces:
a few subspaces dominate the ranking signal, while others contribute marginally.
This deviation suggests that subspace-based retrieval can be both effective and efficient:
by aggregating evidence across multiple subspaces, the true high-scoring keys are more likely to rise to the top, even when each individual subspace provides only partial information.

\subsection{Search path: from key-centric to query-centric}
Prior search-based sparse attention methods only use keys to build key-centric lookup structures during the prefilling stage~\citep{zhang2025pqcache,liu2025clusterkv}.
During decoding, these methods typically follow a \emph{key-centric} search path: \mbox{$Q \!\to\! K$-centroid $\!\to\! K$}.
However, we run Llama-3.1-8B-Instruct on a random NarrativeQA example and visualize $Q$ and $K$ from Layer 10, Head 0 (Figure~\ref{fig:three-across}(d)). The distributions of $Q$ and $K$ diverge significantly.
This misalignment stems from the fact that 
$Q$ and $K$ are generated by different projections, which can be highly anisotropic and vary across layers/heads/timesteps, especially under stylistic/domain shifts. 
Key-only lookup structures (built on $K$) can become out-of-distribution (OOD) for $Q$ to search, causing unstable recall at high sparsity.

A \emph{query-centric} search path, denoted as \mbox{$Q \!\to\! Q$-centroid $\!\to\! K$}, offers greater stability.
Since the nearest-centroid assignment occurs \emph{in the same space} as $Q$, it significantly reduces OOD risk.
Once the query-centric centroid is selected, the search strategy consults only precomputed centroid$\!\to$key partials rather than requiring any online $Q{\to}K$ centroid hop, thereby notably improving recall stability under high sparsity.

\begin{figure*}[t]
\centering
\includegraphics[width=0.65\linewidth]{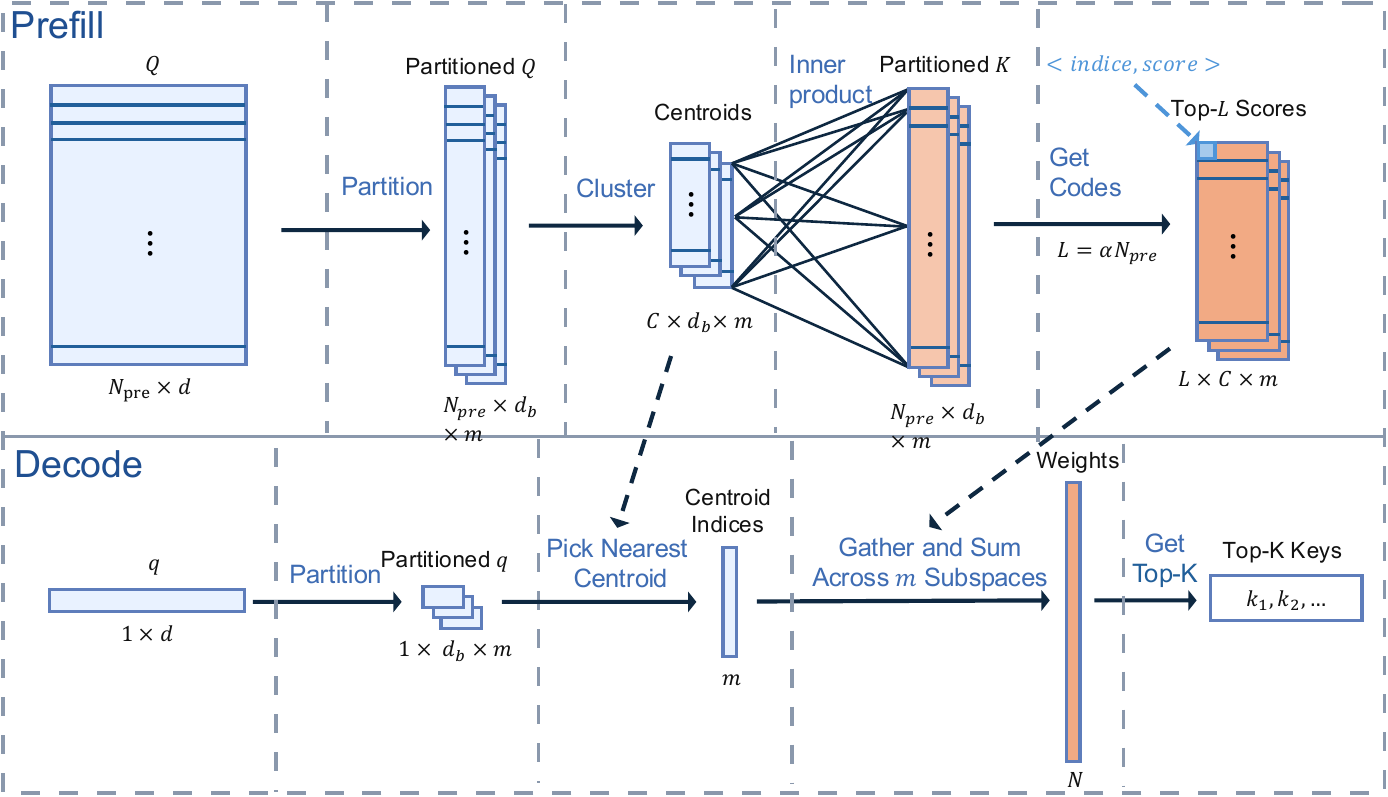}
\caption{\textbf{CSAttention overview.} \emph{Prefill (top):} build query-centric centroids per subspace, score all keys per centroid, and store fixed Top-$L$ $(\text{idx},\text{score})$ lists. \emph{Decode (bottom):} pick nearest centroid in each subspace, fetch $m$ lists, reduce-by-key, then take Top-K and run standard KV gather/attention.}\vspace{-1.0em}
\label{fig:overview}
\end{figure*}

\section{Methodology}

\subsection{Overview of CSAttention}
\label{sec:arch}

\textbf{Architecture overview.}
Figure~\ref{fig:overview} provides an overview of CSAttention, which consists of an offline prefilling stage and an online decoding stage.
CSAttention does not replace the standard KV cache; it augments it with bounded-capacity lookup tables built during prefill. We do not aim to eliminate the KV cache's linear storage; instead, we reduce the amount of KV \emph{touched} per decode step with predictable online work.
To enhance inference efficiency, CSAttention employs a subspace partitioning strategy, as analyzed in Section~\ref{sec:subspace}.
We split $d$ dimensions into $m$ subspaces with sizes $\{d_b\}_{b=1}^m$ and $\sum_b d_b=d$.
For key $k_i \in K=[k_1,k_2,...,k_N]$ and query $q$,
$
k_i=\big(k_i^{(1)},\dots,k_i^{(m)}\big),\quad
q=\big(q^{(1)},\dots,q^{(m)}\big),\quad
q k_i^\top=\sum_{b=1}^m \underbrace{q^{(b)} (k_i^{(b)})^\top}_{\text{subspace partial}}.
$
During prefilling, operations are only performed independently within each subspace: queries are clustered, and the inner products between each centroid and all keys are computed and recorded as \emph{centroid-scores} in a lookup table.
During decoding, a query-centric search is conducted inside each subspace, after which centroid-scores are sparsely accumulated across subspaces to efficiently retrieve the most critical tokens.

\textbf{Design overview.}
(i) Subspace split. We use uniform split by default ($d_b{=}d/m$); nonuniform splits are possible when heads emphasize bands.
(ii) Normalization. We $\ell_2$-normalize subspace vectors when clustering and centroid scoring (cosine scores); the model’s native scaling is preserved for attention.
(iii) Bounded-capacity lookup tables. Each centroid $(b,j)$ stores contiguous arrays
$\mathcal{I}_j^{(b)}\!\in\!\mathbb{N}^{L}$ and $\mathcal{V}_j^{(b)}\!\in\!\mathbb{R}^{L}$, sorted by score.
This guarantees coalesced loads and a bounded decode union ($\le mL$).
The table \emph{capacity} (Top-$L$ per centroid) is fixed after prefill and does not grow with generation length; online updates only replace entries without resizing.


\subsection{Offline (Prefill): Query Clustering and Per-Centroid Scoring}
\label{sec:prefill}

\textbf{(1) Subspace partition and queries clustering.}
Given prefill queries $Q\!\in\!\mathbb{R}^{\PrefillLen \times d}$, partition each $q$ into $\{q^{(b)}\}_{b=1}^m$ and run mini-batch $k$-means \emph{on GPU} per subspace to obtain $C$ centroids
$\{c_j^{(b)}\}_{j=1}^C$:
\[
\min_{\{c_j^{(b)}\}}\ \sum_{q\in Q} \min_{j\in[1..C]} \ \bigl\|\tfrac{q^{(b)}}{\|q^{(b)}\|_2}-c_j^{(b)}\bigr\|_2^2,
\quad \text{s.t.}\ \|c_j^{(b)}\|_2=1.
\]
We use cosine $k$-means (normalize vectors); seeds are $k$-means++ with a small number of iterations.

\textbf{(2) Per-centroid scoring in each subspace.}
For each centroid $(b,j)$ we compute
\[
s_j^{(b)}(i)=c_j^{(b)} (k_i^{(b)})^\top,\qquad i\in[1...\PrefillLen],
\]
via batched GEMM across centroids, where $\PrefillLen$ denotes the prefill sequence length; then keep Top-$L$ pairs $(i,s_j^{(b)}(i))$ and serialize into $(\mathcal{I}_j^{(b)},\mathcal{V}_j^{(b)})$ on the target device (GPU for All-GPU; CPU DRAM for CPU$\leftrightarrow$GPU).
Tables have bounded capacity (Top-$L$ per centroid) and can be reused across requests that share the long prefill.


\subsection{Online (Decode): Query-centric search and Sparse Accumulation}
\label{sec:decode}
Given a new query $q$:

\textbf{(1) Nearest query-centroid per subspace.}
For each subspace $b$,
\[
\hat{j}_b = \arg\max_{j\in[1..C]} \ \tilde{q}^{(b)} (c_j^{(b)})^\top,\quad \tilde{q}^{(b)}=q^{(b)}/\|q^{(b)}\|_2.
\]
Implementation is a batched GEMV over $C$ centroids per subspace (per head), mapping well to GPU.

\textbf{(2) Gather $m$ short lists and build the union.}
Fetch $(\mathcal{I}_{\hat{j}_b}^{(b)},\mathcal{V}_{\hat{j}_b}^{(b)})$ for each $b$ and concatenate to form up to $mL$ $(\text{idx},\text{score})$ pairs.

\textbf{(3) Reduce-by-key (sparse accumulation).}
Let $U$ denote the concatenated indices. We compute the aggregated score per key-id $i\in U$ as
\[
\mathrm{score}(i)=\sum_{b=1}^m w_b\cdot \mathcal{V}_{\hat{j}_b}^{(b)}[i], \quad i\in U,
\]
with $|U|\le mL$. We use uniform subspace weights ($w_b\!=\!1$), learned or confidence-based $w_b$ are possible but not required in our best settings.
This step is branchless and implemented with warp-synchronous reductions. This step realizes \emph{centroid-scoring}: aggregated scores are sums of precomputed centroid$\to$key partials, avoiding any online $Q{\to}K$ code movement.

\textbf{(4) Merge recent window and select Top-K.}
Union the recent window $\{N-R+1,\dots,N\}$ and select Top-K on device. Only these K keys are used in attention; others are ignored.

\textbf{(5) Streaming updates.}
After attention appends a new key $k_S$, we compute 
$\mathrm{score}_{b,j}=(c^{(b)}_j)^\top k^{(b)}_S$ for each subspace $b$ and centroid $j$, 
and try-insert it into the fixed-capacity Top-$L$ list $(\mathcal{I}^{(b)}_j,\mathcal{V}^{(b)}_j)$ (evicting the minimum if needed).
In CPU$\leftrightarrow$GPU mode, this maintenance runs on CPU asynchronously while GPU executes attention.


\subsection{Execution Modes and Memory/Cost Considerations}
\label{sec:modes}

\textbf{CPU$\leftrightarrow$GPU (lookup tables and KV in DRAM; asynchronous execution).}
When HBM capacity is constraining, both the KV cache and the centroid-score lookup tables reside on CPU DRAM, and decoding proceeds with an \emph{explicitly overlapped} CPU–GPU pipeline.
\emph{Prefill:} on the device, Q/K/V projections and full prefill attention run in the forward stream, while per-subspace query clustering and centroid$\!\to$key scoring run in a background stream; the resulting tables are serialized to pinned host memory. Streams are synchronized only once before the first decode step.
\emph{Decode (per step):} (1) the CPU performs the bounded search (select nearest centroids, merge $m$ Top-$L$ lists, and select Top-$K$ by reduce-by-key); (2) transfer only the selected Top-$K$ KV entries to GPU and run attention; (3) CPU performs streaming updates asynchronously by try-inserting new keys into each centroid’s Top-$L$ on the CPU without resizing tables.
\emph{Overlap:} GPU attention at step $t$ overlaps with CPU search and updates for step $t{+}1$. Because only Top-$K$ KV is moved per step, the transfer budget is deterministic and small.

\section{Experiments}
\label{sec:experiments}

\begin{table*}[t]
\centering
\scriptsize
\setlength{\tabcolsep}{2pt}
\caption{LongBench accuracy of sparse methods across three models. Abbreviations: MQA-E (multifieldqa\_en), MQA-Z (multifieldqa\_zh), NarQA (narrativeqa), M-News (multi\_news), Musiq (musique), TrivQA (triviaqa), P-Ret (passage\_retrieval\_en), Hotpot (hotpotqa), G-Rep (gov\_report).}
\resizebox{\linewidth}{!}{%
\begin{tabular}{lccccccccccccccc}
\toprule
 & \multicolumn{15}{c}{\textbf{LongBench Evaluation Tasks}} \\
\cmidrule{2-16}
 & \textbf{MQA-E} & \textbf{MQA-Z} & \textbf{NarQA} & \textbf{M-News} & \textbf{Musiq} & \textbf{Trec} & \textbf{Samsum} & \textbf{TrivQA} & \textbf{P-Ret} & \textbf{Hotpot} & \textbf{G-Rep} & \textbf{LCC} & \textbf{LSHT} & \textbf{VCSum} & \textbf{Avg} \\
\midrule
\multicolumn{16}{c}{\textbf{Llama-3.1-8B-Instruct}} \\
\midrule
Full        & 55.54 & 62.87 & 29.91 & 27.16 & 30.89 & 72.50 & 43.75 & 91.65 & 100.0 & 56.16 & 35.26 & 64.89 & 46.00 & 17.16 & 52.41 \\
CSAttention    & \textbf{56.02} & \textbf{62.01} & \textbf{30.46} & \textbf{26.38} & \textbf{31.11} & \textbf{71.50} & \textbf{44.16} & \textbf{91.95} & \textbf{99.50} & \textbf{55.94} & 33.60 & \textbf{63.33} & \textbf{45.00} & \textbf{17.63} & \textbf{52.04} \\
PQCache & 52.96 & 57.57 & 30.14 & 16.67 & 28.69 & 71.00 & 40.01 & 91.82 & 99.00 & 55.22 & \textbf{34.01} & 60.52 & 43.00 & 16.38 & 49.79 \\
H$_2$O    & 40.17 & 40.01 & 29.21 & 23.94 & 28.08 & 62.00 & 41.10 & 90.32 & 97.00 & 53.36 & 28.32 & 57.96 & 23.50 & 16.50 & 45.11 \\
SparQ   & 39.56 & 34.32 & 26.96 & 21.78 & 28.48 & 47.00 & 42.11 & 89.26 & 87.00 & 51.83 & 25.21 & 55.42 & 21.00 & 15.18 & 41.79 \\
MagicPig & 48.78 & 53.16 & 25.86 & 14.50 & 19.20 & 70.00 & 42.00 & 65.05 & 96.00 & 38.50 & 23.39 & 61.11 & 38.00 & 7.83 & 43.10 \\
\midrule
\multicolumn{16}{c}{\textbf{Qwen3-8B}} \\
\midrule
Full    & 53.67 & 63.37 & 26.05 & 24.88 & 36.18 & 71.50 & 44.30 & 88.54 & 100.0 & 59.40 & 33.35 & 69.13 & 47.50 & 14.31 & 52.30 \\
CSAttention     & \textbf{53.21} & \textbf{63.74} & 26.18 & \textbf{24.60} & \textbf{37.05} & \textbf{72.00} & \textbf{45.10} & \textbf{88.62} & \textbf{100.0} & \textbf{59.18} & 32.63 & \textbf{68.89} & \textbf{46.00} & 14.29 & \textbf{52.25} \\
PQCache & 51.98 & 60.35 & \textbf{26.98} & 21.90 & 36.90 & \textbf{72.00} & 42.80 & 84.10 & \textbf{100.0} & 58.73 & \textbf{33.09} & 61.01 & 44.00 & \textbf{14.39} & 50.59 \\
H$_2$O     & 50.10 & 57.90 & 26.01 & 23.99 & 34.20 & 61.00 & 44.50 & 85.00 & 98.50 & 54.12 & 29.99 & 63.01 & 31.00 & 13.21 & 48.04 \\
SparQ   & 45.32 & 50.07 & 25.88 & 21.03 & 31.90 & 59.00 & 43.90 & 80.20 & 91.00 & 47.93 & 27.62 & 48.32 & 25.50 & 13.48 & 43.65 \\
MagicPig & 52.11 & 57.32 & 26.31 & 18.94 & 28.08 & 58.00 & 44.22 & 87.90 & 98.50 & 51.88 & 24.32 & 55.67 & 40.00 & 9.12 & 46.60 \\
\midrule
\multicolumn{16}{c}{\textbf{Mistral-7B-Instruct-v0.3}} \\
\midrule
Full    & 50.21 & 53.19 & 27.74 & 26.57 & 26.50 & 70.00 & 46.30 & 89.04 & 97.00 & 51.08 & 34.22 & 64.32 & 47.00 & 15.68 & 49.92 \\
CSAttention     & \textbf{49.92} & \textbf{52.94} & 25.56 & \textbf{27.06} & \textbf{26.10} & 70.50 & \textbf{45.91} & \textbf{90.59} & \textbf{97.00} & \textbf{49.34} & \textbf{32.88} & 63.98 & \textbf{46.00} & \textbf{16.44} & \textbf{49.92} \\
PQCache & 45.57 & 39.59 & 22.57 & 26.04 & 22.30 & \textbf{71.00} & 42.18 & 88.62 & 89.00 & 35.22 & 29.68 & \textbf{64.01} & \textbf{46.00} & 15.01 & 45.49 \\
H$_2$O     & 37.26 & 30.43 & 21.07 & 25.33 & 17.01 & 63.00 & 41.98 & 84.77 & 52.00 & 31.56 & 22.92 & 59.91 & 31.00 & 6.04 & 37.45 \\
SparQ   & 31.51 & 31.77 & 19.62 & 21.86 & 15.63 & 61.00 & 41.68 & 84.10 & 42.00 & 29.69 & 25.67 & 53.01 & 34.00 & 5.81 & 35.53 \\
MagicPig & 45.87 & 38.91 & \textbf{26.01} & 23.34 & 21.42 & 71.00 & 45.02 & 90.15 & 95.00 & 34.98 & 31.29 & 55.04 & 29.00 & 14.08 & 44.37 \\
\bottomrule
\end{tabular}%
}
\label{tab:longbench_abbreviated}
\end{table*}

\subsection{Setting}

\paragraph{Models \& baselines.}
We evaluate on three instruction-tuned backbones: Llama3.1-8B, Qwen3-8B, and Mistral-7B (Instruct v0.3).
Baselines include MagicPig (LSH sampling; $L{=}300$, $K{=}10$), SparQ Attention (bandwidth-aware fetching), H\textsubscript{2}O (heavy-hitter retention) using official recommended parameters, alongside PQCache (PQ-based KV retrieval; we give it 15 $k$-means iterations and \texttt{SUBBITS}$=8$ to favor accuracy at high sparsity).
Unless stated otherwise, all methods target a comparable keep ratio near $5\%$.

\paragraph{Hardware.}
Unless otherwise noted, experiments run on a single-node server with dual-socket AMD EPYC 7513 and 1.0\,TiB system memory.  
We bind inference to 64 CPU cores.  
For GPU, we report two regimes: 1$\times$~NVIDIA A100 (single-GPU results) and 4$\times$~NVIDIA A100 on the same host (multi-GPU throughput).  
All methods (ours and baselines) are executed under the same software stack and runtime configuration; identical hardware is used across comparisons.

\paragraph{Datasets}
We use LongBench and LongBench v2. 
LongBench covers 14 datasets across six task categories (single-/multi-doc QA, summarization, few-shot, synthetic, code), with average lengths around 6.7k words (EN) and 13.4k characters (ZH). 
LongBench v2 expands the task set and context range (from $\sim$8k up to the ultra-long regime), emphasizing realistic multi-task retrieval and reasoning. 
We follow official protocols and task metrics (e.g., EM/F1/Acc for QA, ROUGE for summarization) and report per-task and macro-averaged scores.

\paragraph{CSAttention }
Unless otherwise stated we use $m{=}8$ subspaces, $C\!\in\!\{64,128,200\}$ query centroids per subspace, unit subspace weights $w_b{=}1$, and keep $\sim\!5\%$ tokens per step (final Top-K).
We choose $L$ so that $mL$ saturates recall while keeping the reduce-by-key bounded on device; subspace $k$-means uses 10 iterations on GPU (cosine $k$-means with k-means++ seeding).
We evaluate both execution backends: All-GPU (tables+KV resident on GPU) and CPU$\leftrightarrow$GPU (search/gather on CPU, transfer only Top-K KV to GPU).

\begin{table}[t]
\centering
\small
\caption{LongBench v2 evaluation results on Llama-3.1-8B.}
\resizebox{\linewidth}{!}{%
\begin{tabular}{@{}lcccccc@{}}
\toprule
Method & Overall & Easy & Hard & Short & Medium & Long \\
\midrule
Full & 31.0 & 35.4 & 28.3 & 37.2 & 26.0 & 30.6 \\
CSAttention & \textbf{31.2} & \textbf{34.4} & \textbf{29.3} & \textbf{37.8} & 25.1 & \textbf{32.4} \\
PQCache & 29.8 & 33.3 & 27.7 & \textbf{37.8} & 22.3 & 31.5 \\
H$_2$O & 29.9 & 32.9 & 28.0 & 33.8 & 27.9 & 31.5 \\
SparQ & 26.2 & 27.6 & 25.4 & 30.0 & 22.3 & 27.8 \\
MagicPig & 29.2 & 29.5 & 29.0 & 31.8 & \textbf{26.9} & 29.4 \\
\bottomrule
\end{tabular}
}
\label{tab:longbench_v2}
\end{table}


\begin{table*}[t]
\centering
\small
\caption{Task-level accuracy on LongBench for CSAttention schedules (Llama3 8B). 
Schedules are denoted as “keep ratio + search period”: \emph{0.05-step-1} keeps 5\% tokens and searches every step; \emph{0.15-step-4} keeps 15\% and searches every 4 steps; etc. 
All schedules are accuracy-stable relative to Full; macro-average gaps are \(\le 0.6\) points.}
\setlength{\tabcolsep}{4.5pt}
\resizebox{\linewidth}{!}{%
\begin{tabular}{lccccccccccccccc}
\toprule
 & \textbf{MQA-E} & \textbf{MQA-Z} & \textbf{NarQA} & \textbf{M-News} & \textbf{Musique} & \textbf{Trec} & \textbf{Samsum} & \textbf{TrivQA} & \textbf{P-Ret} & \textbf{Hotpot} & \textbf{G-Rep} & \textbf{LCC} & \textbf{LSHT} & \textbf{VCSum} & \textbf{Avg} \\
\midrule
Full                 & 55.54 & 62.87 & 29.91 & 27.16 & 30.89 & 72.50 & 43.75 & 91.65 & 100.00 & 56.16 & 35.26 & 64.89 & 46.00 & 17.16 & 52.41 \\
\midrule
\textbf{0.05 step 1} & \textbf{56.02} & 62.01 & \textbf{30.46} & 26.38 & 31.11 & 71.50 & 44.16 & \textbf{91.95} & \textbf{99.50} & \textbf{55.94} & 33.60 & \textbf{63.33} & 45.00 & \textbf{17.63} & \textbf{52.04} \\
\textbf{0.05 step 2} & 53.98 & 60.91 & 28.99 & 26.05 & 30.09 & 70.50 & 42.18 & 90.05 & 97.50 & 54.03 & 31.56 & 63.21 & 45.00 & 17.32 & 50.81 \\
\textbf{0.15 step 4} & 54.97 & 62.04 & 30.28 & \textbf{26.95} & 30.72 & 71.00 & \textbf{44.43} & 91.89 & 99.00 & 55.64 & \textbf{33.84} & 63.12 & 44.50 & 16.78 & 51.80 \\
\textbf{0.20 step 8} & 54.99 & \textbf{62.50} & 29.84 & 26.41 & \textbf{31.43} & \textbf{72.00} & 43.98 & 91.87 & 98.50 & 55.51 & 33.57 & 62.88 & \textbf{45.50} & 17.23 & 51.87 \\
\bottomrule
\end{tabular}%
}
\label{tab:schedule_acc_longbench}
\end{table*}

\subsection{Performance}
\label{sec:acc_results}

\paragraph{Results on LongBench.}
Table~\ref{tab:longbench_abbreviated} reports per-task accuracy on LongBench for three backbones.
On Llama3.1-8B CSAttention’s macro average (52.04) is within $0.7\%$ loss of Full (52.41), with numerous per-task wins (e.g., MQA-E/Z, NarQA, Musique, TrivQA) and near-ties elsewhere.
Qwen3-8B shows virtually identical averages (52.25 vs.\ 52.30), again with CSAttention matching or exceeding Full on multiple tasks (e.g., Musique, TrivQA), and never incurring large degradations on any category.
On Mistral-7B, CSAttention matches the Full average exactly (both 49.92) while leading or tying on several tasks (e.g., TrivQA, P-Ret, VCSum), indicating robustness across architectures.
In contrast, PQCache—despite being tuned with 15 $k$-means iterations and \texttt{SUBBITS}$=8$—and H$_2$O/SparQ/MagicPig all trail CSAttention on the macro average and drop notably on harder retrieval/summarization tasks (e.g., M-News, Hotpot), consistent with their sensitivity to high sparsity.

\paragraph{Results on LongBench v2.}
Table~\ref{tab:longbench_v2} presents the LongBench v2 accuracy results for Llama-3.1-8B.
CSAttention achieves an overall score of 31.2, surpassing the dense Full attention baseline (31.0) and exceeding all sparse baselines.
Notably, while keeping only $\sim$5\% tokens, CSAttention maintains performance within statistical noise of the Full model on the global metric, whereas other sparse methods (PQCache, H$_2$O, SparQ, and MagicPig) exhibit more substantial drops (ranging from $-1.2$ to $-4.8$ points overall).
These results demonstrate that our centroid-scoring token retrieval mechanism effectively preserves \emph{near-full} model accuracy even at very high sparsity levels.
Furthermore, CSAttention maintains strong performance on Easy and Short tasks (surpassing other sparse methods) while demonstrating particular strength on challenging Hard and Long tasks. The method improves Hard performance from 28.3 (Full) to 29.3 and Long from 30.6 (Full) to 32.4, suggesting enhanced capability for complex, long-context tasks.

\paragraph{Stability at 95\% sparsity.}
Across all three backbones, the gap between CSAttention and Full on the macro average is $\leq 0.37$ points, and exactly zero for Mistral-7B.
Moreover, CSAttention’s per-task variance is modest: it avoids catastrophic failures observed in some baselines (e.g., pronounced declines on M-News or cross-lingual QA).
Combined with the length-bucket analysis, these results support the claim that \emph{subspace partition + centroid-scoring in $Q$-space} maintains high recall of truly relevant keys under very high sparsity, delivering accuracy that is \emph{indistinguishable from Full} in practice.

\paragraph{Schedules evaluation.}
Table~\ref{tab:schedule_acc_longbench} reports task-level accuracy on LongBench for several CSAttention schedules under identical model and data settings. 
Three observations emerge. 
\emph{(i) Near-full accuracy at 95\% sparsity.} \emph{0.05-step-1} (keep 5\%, search every step) attains an average of 52.04, within \(\mathbf{0.37}\) points of Full (52.41), and tracks Full closely across QA, summarization, and retrieval tasks. 
\emph{(ii) Infrequent search preserves accuracy.} Leveraging the empirical locality that consecutive tokens tend to share similar attention patterns, both \emph{0.15-step-4} (keep 15\%, search every 4 steps; 51.80) and \emph{0.20-step-8} (keep 20\%, search every 8 steps; 51.87) remain within \(\le\!0.61\) points of Full on the macro average, with no catastrophic drops on any task. 
\emph{(iii) Robustness vs.\ PQ-style retrieval.} Even at the most aggressive sparsity, \emph{0.05-step-1} matches or exceeds the accuracy typically observed for PQCache-like methods at comparable keep budgets (cf. Section~\ref{sec:acc_results}), reflecting the stability of subspace partition and query-space centroid-scoring.

\begin{figure}[t]
  \centering
   \includegraphics[width=0.95\linewidth]{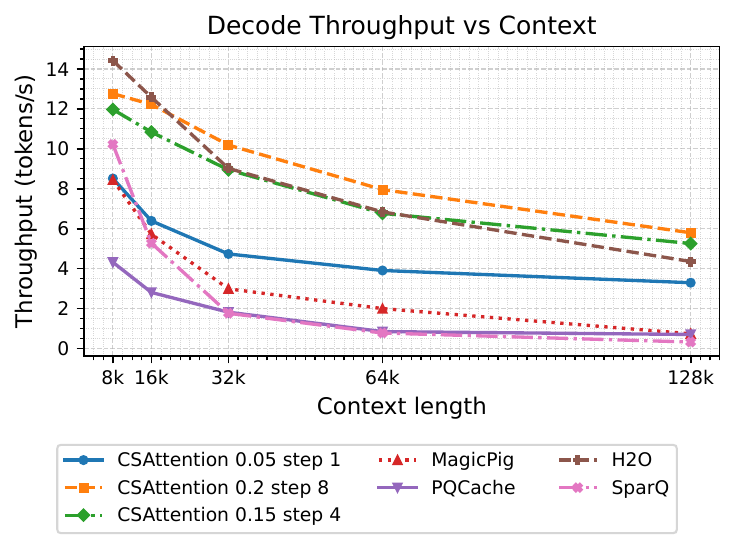}
  \caption{\textbf{Long-context decode efficiency.}}
  \label{fig:cpu_gpu_decode}
\end{figure}

\subsection{Efficiency}
\label{sec:efficiency}

\paragraph{Schedules.}
We report three CSAttention schedules that trade sparsity and search frequency while maintaining \emph{near-full} accuracy (Table~\ref{tab:schedule_acc_longbench}):
\emph{0.05-step-1} keeps $5\%$ tokens (95\% sparsity) and searches every step;
\emph{0.20-step-8} keeps $20\%$ tokens and searches every $8$ steps;
\emph{0.15-step-4} keeps $15\%$ tokens and searches every $4$ steps.

\paragraph{Decode throughput under CPU$\leftrightarrow$GPU mode.}
As illustrated in Figure~\ref{fig:cpu_gpu_decode}, once the index is preloaded, CSAttention attains state-of-the-art decode throughput in the CPU$\leftrightarrow$GPU setting and the advantage \emph{grows with context length}.
Using the best CSAttention schedule at each length, speedups over baselines are:
\begin{itemize}
\item vs.\ PQCache: \textbf{$2.95\times$} (8K), \textbf{$4.35\times$} (16K), \textbf{$5.60\times$} (32K), \textbf{$9.40\times$} (64K), \textbf{$8.26\times$} (128K).
\item vs.\ MagicPig: \textbf{$1.51\times$} (8K), \textbf{$2.14\times$} (16K), \textbf{$3.41\times$} (32K), \textbf{$3.98\times$} (64K), \textbf{$7.85\times$} (128K).
\item vs.\ SparQ: \textbf{$1.25\times$} (8K), \textbf{$2.32\times$} (16K), \textbf{$5.78\times$} (32K), \textbf{$10.3\times$} (64K), \textbf{$17.9\times$} (128K).
\item vs.\ H$_2$O, near parity at short lengths and consistent gains thereafter:
$0.88\times$ (8K), $0.97\times$ (16K), \textbf{$1.13\times$} (32K), \textbf{$1.16\times$} (64K), \textbf{$1.33\times$} (128K).
\end{itemize}
These results validate the intended deployment pattern of \emph{offline prefill + online decode}: a single offline build enables substantially higher online throughput, and the gap widens with longer contexts because CSAttention’s per-step work scales with fixed table sizes rather than total history.


\begin{table}[t]
\centering
\small
\caption{Normalized per-step overhead latency (mean of append+search normalized to 1.00).}
\label{tab:latency_norm_blue}
\begin{tabular}{lccc}
\toprule
\textbf{Schedule} & \textbf{P50} & \textbf{P90} & \textbf{P99} \\
\midrule
\textbf{step 8} & 0.900 & 0.920 & 1.013 \\
\textbf{step 4} & 0.947 & 0.970 & 1.007 \\
\textbf{step 1} & 0.948 & 0.969 & 1.006 \\
\bottomrule
\end{tabular}
\end{table}

\paragraph{Step-level overhead and predictability.}
To complement the throughput curves above, we report step-level overhead statistics for the two components affected by CSAttention (\emph{search} and \emph{append}; the attention kernel is unchanged and thus omitted here). For each schedule, we normalize the mean of (\emph{append}+\emph{search}) to 1.0 and scale P50/P90/P99 accordingly. As shown in Table~\ref{tab:latency_norm_blue}, tails are tight across schedules (P90 $\approx$ 0.92--0.97; P99 $\approx$ 1.01), indicating bounded variability and predictable decode-time behavior. Empirically, the per-step composition is stable: \textbf{Attention\,:\,Search\,:\,Update $\boldsymbol{\approx}$ 1.0\,:\,0.3\,:\,0.1}.
This matches the algorithmic design: per-step work touches at most $mL$ entries during the union--reduce and performs $O(1)$ try-insert for streaming updates, so the overhead remains insensitive to the total history length. These data explain \emph{why} the gains in Figure~\ref{fig:cpu_gpu_decode} strengthen with longer contexts and \emph{why} lower search frequency schedules (e.g., 0.20-step-8) are preferable at short lengths.

\begingroup
\begin{table}[tbp]
\centering
\caption{LongBench-v2 with CoT (max gen 2048): accuracy buckets under prolonged decode. "our" = CSAttention (default schedule).}
\label{tab:r2_cot_final}
\resizebox{\linewidth}{!}{
\begin{tabular}{lcccccc}
\toprule
\textbf{Model} & \textbf{Overall} & \textbf{Easy} & \textbf{Hard} & \textbf{Short} & \textbf{Med.} & \textbf{Long} \\
\midrule
Llama-70B (Full)    & 36.5 & 38.6 & 35.2 & 46.0 & 33.0 & 27.6 \\
Llama-70B (our)     & 36.4 & 39.1 & 34.8 & 44.0 & 35.4 & 25.9 \\
\midrule
Qwen3-32B (Full)   & 49.2 & 53.1 & 46.8 & 60.0 & 41.1 & 47.2 \\
Qwen3-32B (our)    & 48.1 & 51.1 & 46.2 & 57.0 & 40.0 & 49.4 \\
\bottomrule
\end{tabular}
}
\end{table}
\endgroup

\subsection{Robustness under long decode with CoT}
\label{sec:exp_longbench_v2_cot}
\noindent\textbf{Setup and motivation.}
To stress the \emph{long-prefill, long-decode} regime we target, we evaluate CSAttention on LongBench-v2 with \emph{chain-of-thought (CoT)} decoding. We append a simple CoT instruction (e.g., ``think step by step'') to each prompt and allow up to 2048 generated tokens before the final answer. Following LongBench-v2 protocols, only the final answer is scored under the official metrics. Operationally, the long system prompt is prefetched and indexed offline; decode then continues for hundreds or thousands of steps. The question is whether a gradually drifting query distribution during CoT degrades centroid-scoring retrieval.

\medskip
\noindent\textbf{Engineering safeguards for robustness.}
At decode time we use two lightweight mechanisms that keep work \emph{bounded} while improving robustness: (i) a \emph{recent-window candidate union} of size $R$, which unions the most recent $R$ positions into the candidate set before Top-$K$; and (ii) a \emph{centroid backoff} that, when the best cosine similarity is low, selects the top-$\tau$ nearest centroids per subspace ($\tau\!\in\!\{2,3\}$), merging at most $m\tau L$ entries prior to the reduce-by-key. We also allow small, thresholded bumps of $R$ or $K$ in rare “off-manifold’’ cases. All mechanisms preserve fixed-size tables and do not change the asymptotic decode cost.

\medskip
\noindent\textbf{Results.}
In the long-context \emph{offline prefill} setting, for each KV-head group (GQA) we construct centroids by treating the prefill \(\mathbf{Q}\)-embeddings as a distribution over the session/domain. This distribution is already diverse due to the long system prompt, so nearest-centroid routing in $Q$-space remains stable during subsequent long decode. Empirically, Table~\ref{tab:r2_cot_final} shows CSAttention remains \emph{near-baseline} across difficulty buckets for both Llama-70B and Qwen3-32B, indicating that prolonged CoT generation does not materially erode recall at high sparsity in this setting.
\section{Related work}
\label{sec:relatedwork}
Attention serves as the core mechanism in Transformer models~\citep{vaswani2017attention}. 
Standard attention matrices exhibit inherent sparsity, wherein a large fraction of the computed weights are close to zero and can be pruned without significant impact on output quality~\citep{zhang2025efficient}.
By exploiting this sparsity pattern, \emph{sparse attention} methods achieve significant improvements in computational efficiency~\citep{zhangspargeattention,liu2024retrievalattention,desai2024hashattention}.
Based on the mechanism for selecting attention tokens, sparse methods can be categorized into two types: \emph{static methods}, which rely on a predefined sparsity pattern based on empirical observations to fix the computational tokens~\citep{xiao2024efficient,fu2025moa,zhu2024sampleattention,xiaoduoattention}, and \emph{dynamic methods}, which adaptively determine these tokens during decoding according to the real-time distribution of queries and keys~\citep{zhang2023h2o,xiao2024infllm,jiang2024minference,ribar2024sparq,tang2024quest,chen2024magicpig,zhang2025pqcache,singhania2024loki}.
While static methods offer straightforward implementation, their fixed token selection patterns may lead to limitations in capturing long-range dependencies, as well as the potential loss of critical intermediate information~\citep{hu2025raas,tang2024quest}.

Dynamic sparse methods have attracted much attention due to their flexibility and adaptability.
Quest~\citep{tang2024quest} and InfLLM~\citep{xiao2024infllm} adopt a similar strategy: they partition the KV cache into blocks and generate a representative key vector for each block to facilitate efficient searching.
SparQ~\citep{ribar2024sparq} and Loki~\citep{singhania2024loki} estimate the Top-K most relevant keys for a given query by performing dimensionality reduction.
H$_2$O~\citep{zhang2023h2o} maintains a fixed-size KV cache during decoding by dynamically evicting tokens.
MagicPig~\citep{chen2024magicpig}, RetrievalAttention~\citep{liu2024retrievalattention}, HashAttention~\citep{desai2024hashattention}, and PQCache~\citep{zhang2025pqcache} adopt vector search techniques—such as learning to hash, locality-sensitive hashing, and graph—to efficiently retrieve critical tokens.
Our proposed CSAttention also falls into the category of dynamic sparse methods, exhibiting superior efficiency and effectiveness in LLM inference compared to existing techniques.

\section{Conclusion}
\label{sec:conclusion}

In this paper, we introduced Centroid-Scoring Attention (CSAttention), a training-free sparse attention method for efficient LLM inference.
CSAttention ensures the reliable recovery of high-scoring keys under very high sparsity by mitigating the query-key distribution shift through subspace partitioning and query-centric table construction.
Extensive experiments demonstrate that compared to state-of-the-art sparse attention methods, CSAttention maintains near-lossless model accuracy while achieving higher inference speed in high-sparsity and long-context scenarios, demonstrating its practical value for scalable long-context inference.

\bibliography{ref}
\bibliographystyle{icml2026}

\clearpage
\appendix


\section{All-GPU (lookup tables and KV on HBM).}
\label{sec:all-gpu}
When HBM is sufficient, we keep both tables and KV on-device.
Decode is: nearest-centroid (batched GEMV) $\rightarrow$ coalesced gather of $m$ Top-$L$ lists $\rightarrow$ reduce-by-key (radix sort + segmented sum) $\rightarrow$ device Top-$K$ $\rightarrow$ standard attention over selected KV. This mode maximizes throughput and avoids PCIe transfers, while preserving the same bounded-work per-step structure.

\section{Additional experimental details and results}
\label{sec:app_results}

\subsection{Schedules evaluation}
\label{sec:schedules}

\paragraph{Efficiency.}
As shown in Figure~\ref{fig:decode_two_panels}, at 8K–16K context lengths, the lower search frequency of \emph{0.20-step-8} yields the highest throughput. 
Beyond 32K, however, \emph{0.15-step-4} and \emph{0.05-step-1} become preferable as their stronger recall translates into better final throughput under the same token retention budget.
Under the CPU\(\leftrightarrow\)GPU mode (cf. Figure~\ref{fig:cpu_gpu_decode}), less frequent searching reduces control traffic and amortizes host-side list merging along with PCIe H2D transfers of gathered KV (only the selected \(\text{K}{=}\rho \SeqLen\) vectors are moved). 
Consequently, schedules such as \emph{0.15-step-4} and \emph{0.20-step-8} deliver higher throughput at short to medium lengths, whereas \emph{0.05-step-1} dominates at very long contexts by maximizing recall under a fixed token keep rate.
Notably, all CSAttention scheduling variants maintain \emph{near-lossless} accuracy compared to Full attention (cf. Table~\ref{tab:schedule_acc_longbench}), allowing schedule selection to be driven primarily by throughput considerations for any given context length requirement.



\subsection{Prefill latency.}

\begin{figure}[t]
  \centering
   \includegraphics[width=0.95\linewidth]{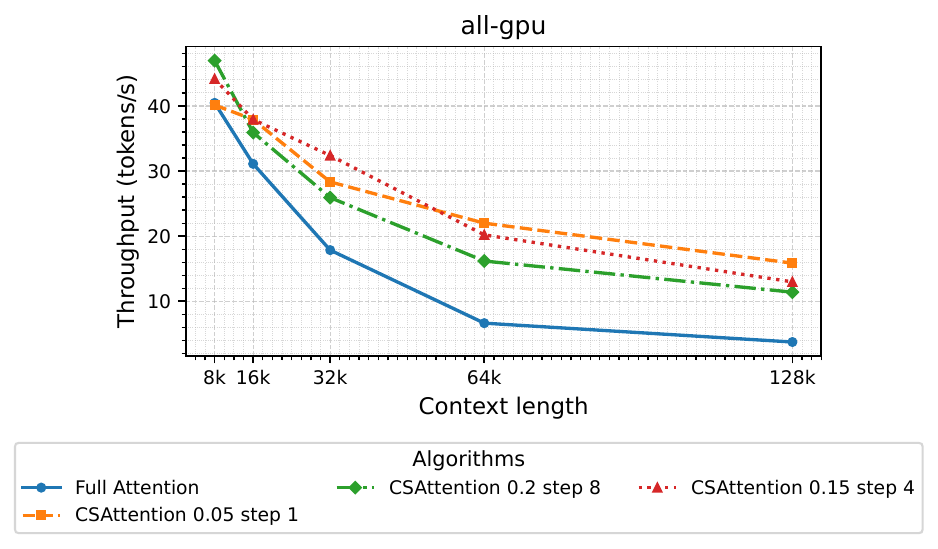}
  \caption{\textbf{All-GPU mode decode efficiency.}}
  \label{fig:decode_two_panels}
\end{figure}

\begin{figure*}[t]
  \centering
  \includegraphics[width=0.9\linewidth]{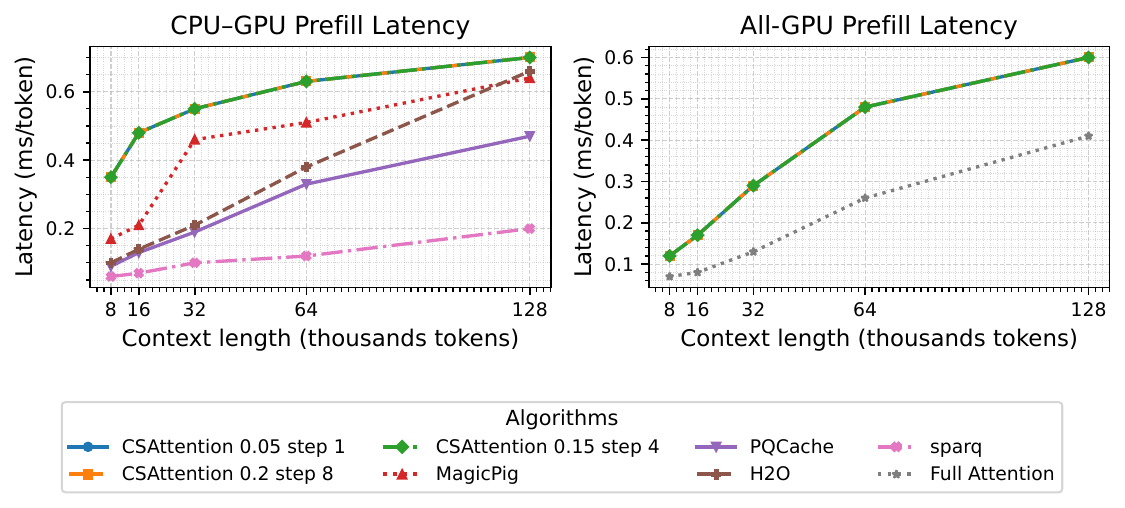}
  \caption{\textbf{Long-context prefill latency.}
  Left: CPU$\leftrightarrow$GPU mode; Right: All-GPU mode.}
  \label{fig:prefill_two_panels}
\end{figure*}

As illustrated in Figure~\ref{fig:prefill_two_panels}, CSAttention invests additional computation during the prefill stage to construct its query-centric tables. 
This upfront cost, which is \emph{offline} and incurred once per shared prompt and amortized over subsequent decoding steps, is a deliberate design trade-off to achieve the significant acceleration and robustness observed during online decoding (cf. Figure~\ref{fig:decode_two_panels}).

\section{Complexity analysis}
\label{sec:app_complexity}

\paragraph{Notation (recap).}
We use column vectors and $\langle\cdot,\cdot\rangle$ for inner products.
Hidden size per head is $d$, partitioned into $m$ subspaces with sizes $\{d_b\}_{b=1}^m$ and $\sum_{b=1}^m d_b=d$.
We denote the fixed prefill sequence length as $\PrefillLen$ and the growing sequence length as $\SeqLen$ (number of cached keys/values so far).
Each subspace has $C$ centroids; for centroid $(b,j)$ the offline list length is fixed based on the prefill size:
\[
L \;=\; \alpha \PrefillLen,\quad \alpha\in(0,1).
\]
At decode we keep $\text{K}=\rho \SeqLen$ keys ($\rho\in(0,1)$).
Tables are $(\mathcal{I}_j^{(b)},\mathcal{V}_j^{(b)})$ with indices in \texttt{int32} (4B) and scores in \texttt{fp16}/\texttt{bf16} (2B).
Unless otherwise stated, complexities are \emph{per head, per layer}; extension across layers/heads is linear. We denote by $I$ the number of mini-batch $k$-means iterations used for subspace clustering during prefill (we use $I{=}10$ by default on GPU).

\paragraph{All-GPU complexity.}
Per query (per head):
\[
\begin{aligned}
&\text{Nearest centroid}: && O(C d),\\
&\text{Concat/sort/sum over } U: && O(m\alpha \PrefillLen),\\
&\text{Device Top-}\text{K} \text{ on }U: && O(m\alpha \PrefillLen),\\
&\text{Sparse attention on }\text{K}=\rho \SeqLen: && O(\rho \SeqLen\, d).
\end{aligned}
\]
Hence total time is:
\[
T_{\mathrm{all\text{-}GPU}}(\SeqLen)= \underbrace{O(Cd) + O(m\alpha \PrefillLen)}_{\text{Constant Search Cost}} \;+\; \underbrace{O(\rho \SeqLen\, d)}_{\text{Linear Attention}}.
\]
Unlike dense attention $O(\SeqLen d)$, CSAttention's retrieval overhead does not grow with generation length.

\paragraph{CPU$\leftrightarrow$GPU complexity and bandwidth (asynchronous).}
On the host, the bounded search is a linear merge/reduce over a fixed number of elements:
\[
T_{\text{CPU-search}} \;=\; O(m\alpha \PrefillLen).
\]
Gather moves only the selected $\text{K}=\rho \SeqLen$ keys/values to device. H2D bytes per step scale linearly:
\[
\mathrm{H2D\_bytes/step} \;=\; 2\,\rho \SeqLen \cdot d \cdot \mathrm{B}.
\]
With search period $P>1$ (reuse the index set for $P{-}1$ steps), the amortized host work and H2D shrink by $\approx 1/P$:
\[
\overline{T}_{\text{CPU-search}} \approx O\!\Big(\tfrac{m\alpha}{P} \PrefillLen\Big), \quad
\overline{\mathrm{H2D\_bytes/step}} \approx \tfrac{1}{P}\cdot 2\,\rho \SeqLen d\,\mathrm{B}.
\]
Device-side attention remains $O(\rho \SeqLen d)$.
Because CPU search + H2D for step $t{+}1$ overlap with GPU attention at step $t$ (streams/events, pinned buffers), PCIe latency is largely hidden; wall time is dominated by $O(\rho \SeqLen d)$.

\subsection{Prefill}

\paragraph{KV cache baseline.}
For one head in one layer, dense KV storage for length $\SeqLen$ is
\[
\mathrm{KV\_bytes} \;=\; 2\,d\,\mathrm{B}\cdot \SeqLen,
\]
where the factor 2 accounts for K and V, and $\mathrm{B}\!\in\!\{2,4\}$ is bytes per element (fp16/bf16 or fp32).
Across layers the footprint scales linearly with $\SeqLen$ and dominates device memory at long context.

\paragraph{Index (tables) footprint (Fixed).}
Per head, our lists store $mC L$ entries. Since $L$ is determined by the prefill length ($L=\alpha \PrefillLen$), the index size is \textbf{constant} during decoding:
\[
\underbrace{m\,C\,(\alpha \PrefillLen)\cdot 6}_{\text{lists (fixed capacity)}} \;+\;
\underbrace{m\,C\,d_b\cdot 2}_{\text{centroids (negligible)}} \quad\text{bytes.}
\]
Comparing the table bytes to KV bytes \emph{at the end of prefill} (when $\SeqLen = \PrefillLen$):

\[
\frac{\text{table bytes}}{\text{KV bytes}}
\;\approx\;
\frac{6\,mC\alpha}{2\,d\,\mathrm{B}}
\;=\;
\begin{cases}
\dfrac{3\,mC\alpha}{2\,d}, & \mathrm{B}=2~(\text{fp16/bf16}),\\[6pt]
\dfrac{3\,mC\alpha}{4\,d}, & \mathrm{B}=4~(\text{fp32}).
\end{cases}
\]

Crucially, as decoding proceeds and $\SeqLen$ grows ($\SeqLen \gg \PrefillLen$), the KV cache grows linearly while the table remains fixed. Thus, the \textbf{relative memory overhead of CSAttention decreases} asymptotically toward zero.

\paragraph{Prefill time.}
Per subspace, centroid$\to$key scoring multiplies $K^{(b)}\!\in\!\mathbb{R}^{\PrefillLen\times d_b}$ with $(C^{(b)})^\top\!\in\!\mathbb{R}^{d_b\times C}$, costing $O(\PrefillLen C d_b)$.
Summed over $m$:
\[
T_{\text{prefill}} \;=\;
O\!\big(I\,\PrefillLen C d\big) \;+\; O\!\big(\PrefillLen C d\big) \;+\; O\!\big(\PrefillLen C m\big),
\]
dominated by subspace $k$-means ($I$ iterations). 
Prefill is a one-time cost per shared prompt and is amortized across requests.

\subsection{Decode}

Let $U$ be the union of indices retrieved from the $m$ lists at a step. Since $L$ is fixed ($L=\alpha \PrefillLen$):
\[
|U| \;\le\; mL \;=\; m\alpha \PrefillLen \quad (\text{Constant w.r.t. } \SeqLen).
\]

\paragraph{Why the union remains effective.}
Although the potential search space is large, our union $|U|\le m\alpha \PrefillLen$ is strictly bounded by the prefill size.
The accumulator exploits $\langle {q},{k}_i\rangle=\sum_{b=1}^m \langle {q}^{(b)},{k}_i^{(b)}\rangle$.
Truly aligned keys “collide” across subspaces and rise in the index-wise sum. This property ensures that a moderate, fixed $\alpha$ (relative to prefill) suffices to saturate recall even as the sequence grows, keeping the \emph{search-side} cost constant while the \emph{attention-side} coefficient $\rho$ controls the growing compute.

\paragraph{Space/time takeaway under long context.}
Online memory is dominated by the KV cache $O(\SeqLen d)$ which grows linearly; CSAttention adds lists of size $O(mC\alpha \PrefillLen)$ which remains constant regardless of generation length.
Per-step time replaces the heavy $O(\SeqLen d)$ term with a smaller attention term plus a constant search term:
\[
O(\SeqLen d)
\Rightarrow
\underbrace{O\big(m\alpha \PrefillLen\big)}_{\text{Constant Search}} + \underbrace{O\big(\rho \SeqLen d\big)}_{\text{Sparse Attn}} \;+\,O(Cd).
\]
With typical settings (e.g., $\rho\!\approx\!0.05$, $\alpha\!\in\![0.1,0.4]$), the decode cost and H2D traffic are substantially smaller than dense, and the one-time prefill/index overhead is amortized across many decode steps and many requests that share the prefill.

\paragraph{Decode throughput under All-GPU mode.}
Figure~\ref{fig:decode_two_panels} (Right) compares Full attention with the three CSAttention schedules under an all-GPU backend.
CSAttention consistently outperforms Full attention across all context lengths, with the performance advantage emerging early and growing substantially as sequence length increases: $1.16\times$ speedup at 8K, $1.22\times$ speedup at 16K, $1.81\times$ speedup at 32K, $3.31\times$ speedup at 64K, and $4.24\times$ speedup at 128K.
The performance gains arise from replacing $O(N)$ dense inner products with \emph{fixed-size} list lookups and device-side union–reduce–Top-K kernels (cf. Section~\ref{sec:decode}), whose cost is insensitive to history length.

\begin{table}[tbp]
\centering
\caption{\textbf{Recommended defaults (per KV head) and tuning guidance.}}
\label{tab:defaults_final}
\resizebox{\linewidth}{!}{%
\begin{tabular}{@{}ll@{}}
\toprule
\textbf{Parameter} & \textbf{Default / Guidance} \\
\midrule
Subspaces $m$ & $8$ (use $6$ if memory-bound; $12$ for ultra-long contexts) \\
Centroids $C$ & $64$ ($32$ for compact tables; $128$ for very long $N$) \\
Top-$L$ & $L{=}\alpha \PrefillLen$, $\alpha{=}0.2$ (tune $0.10{\sim}0.40$) \\
Keep $K$ & $K{=}\rho N$, $\rho{=}0.05$ (95\% sparsity) \\
Recent window $R$ & $32$ (increase to $128$ for stronger recency bias) \\
Subspace weights $w_b$ & $1$ (learned/confidence weights optional) \\
\bottomrule
\end{tabular}
}
\end{table}

\section{Hyperparameters \& Index Sizing}
\label{app:hyperparams}

\paragraph{Scope.}
This appendix provides recommended defaults, tuning ranges, and sizing guidance for CSAttention. Unless noted otherwise, parameters are specified \emph{per KV head}.

\paragraph{Recommended defaults.}
We recommend: subspaces $m{=}8$; centroids per subspace $C\!\in\!\{32,64,128\}$ (default $64$; use $32$ for tighter memory, $128$ for ultra-long contexts); Top-$L$ uses proportional scaling $L=\alpha \PrefillLen$ (default $\alpha{=}0.20$, tunable $0.10{\sim}0.40$); keep ratio $\rho{=}0.05$ (Top-$K=\rho N$); uniform subspace weights $w_b{=}1$; recent passthrough window $R\in[16,128]$ (default $32$).

\paragraph{Sensitivity (what matters).}
The decode latency consists of a fixed search cost (determined by $mL$) and a growing attention cost (determined by $K$). In practice:
\begin{itemize}
  \item $m\in\{6,8,12\}$: Increasing $m$ improves multi-subspace “collision” recall but grows table size and reduction work linearly; $m{=}8$ is a robust knee.
  \item $C\in\{32,64,128\}$: Larger $C$ improves centroid coverage, but returns diminish because decode selects only the nearest centroid per subspace; $C{=}64$ is a strong default.
  \item $\alpha$ controls the fixed list size $L{=}\alpha \PrefillLen$: higher $\alpha$ boosts recall but increases the constant union--reduce overhead; $\alpha{=}0.2$ balances accuracy and base latency.
  \item $\rho$ sets the final Top-$K$ density; it linearly affects attention FLOPs and (in CPU$\leftrightarrow$GPU mode) PCIe bytes per step. $\rho{=}0.05$ achieved near-lossless accuracy in our runs.
\end{itemize}


\paragraph{Practical presets.}
For medium lengths (8–32K) emphasizing throughput: $m{=}8$, $C{=}64$, $\alpha{=}0.20$, $\rho{=}0.05$. For ultra-long lengths (64–128K) emphasizing near-lossless accuracy: $m{=}8$, $C{=}128$, $\alpha{=}0.1$, $\rho{=}0.05$.




\end{document}